# Time Series Analysis and Forecasting of COVID-19 Cases Using LSTM and ARIMA Models

Arko Barman, Ph.D.[1]

**KEYWORDS:** COVID-19, Time Series Analysis, Forecasting, LSTM, ARIMA

## ABSTRACT

**Objective**

Coronavirus disease 2019 (COVID-19) is a global public health crisis that has been declared a pandemic by World Health Organization. Forecasting country-wise COVID-19 cases is necessary to help policymakers and healthcare providers prepare for the future. This study explores the performance of several Long Short-Term Memory (LSTM) models and Auto-Regressive Integrated Moving Average (ARIMA) model in forecasting the number of confirmed COVID-19 cases.

**Materials and Methods**

Time series of daily cumulative COVID-19 cases were used for generating 1-day, 3-day, and 5-day forecasts using several LSTM models and ARIMA. Two novel k-period performance metrics – k-day Mean Absolute Percentage Error (kMAPE) and k-day Median Symmetric Accuracy (kMdSA) – were developed for evaluating the performance of the models in forecasting time series values for multiple days.

**Results**

Errors in prediction using kMAPE and kMdSA for LSTM models were both as low as 0.05%, while those for ARIMA were 0.07% and 0.06% respectively. LSTM models slightly underestimated while ARIMA slightly overestimated the numbers in the forecasts.

**Discussion**

The performance of LSTM models is comparable to ARIMA in forecasting COVID-19 cases. While ARIMA requires longer sequences, LSTMs can perform reasonably well with sequence sizes as small as 3. However, LSTMs require a large number of training samples. Further, the development of k-period performance metrics proposed is likely to be useful for performance evaluation of time series models in predicting multiple periods.

**Conclusion**

Based on the k-period performance metrics proposed, both LSTMs and ARIMA are useful for time series analysis and forecasting for COVID-19.

---

[1] Arko.barman@uth.tmc.edu

# 1 OBJECTIVE

The novel coronavirus (nCOV-19) and the associated disease, coronavirus disease 2019 (COVID-19), presumably originated from Wuhan, China in December 2019[1]. The rapid spread of COVID-19 worldwide caused the World Health Organization (WHO) to declare it as a pandemic in March 2020[1]. As of May 25, 2020, more than 1.5 million people in the United States and nearly 6 million people worldwide have been confirmed with COVID-19[2]. Compared to other highly-contagious previously-identified coronavirus-related diseases, such as Severe Acute Respiratory Syndrome (SARS) and Middle East Respiratory Syndrome (MERS), COVID-19 appears to be more contagious according to preliminary studies[3].

In light of the present circumstances, it is of paramount importance for policymakers and healthcare professionals to plan for the future and be adequately equipped for circumstances that may arise due to the rapid spread of COVID-19. A crucial part of planning ahead in this scenario is to forecast the number of cases in the future, both in the short term and in the long term. While long term predictions and forecasting can be made with epidemiological models, short term predictions can be performed by time series analysis. In this study, we evaluate the performance of several time series analysis and forecasting models for predicting country-wise COVID-19 cases in the short term.

# 2 BACKGROUND AND SIGNIFICANCE

For forecasting using time series analysis, several models, such as Long Short-Term Memory (LSTM) networks[4] and Auto-regressive Integrated Moving Average (ARIMA) model[5], have been widely used. Some preliminary studies for COVID-19 time series forecasting using ARIMA have also been done[6,7,8]. However, two avenues of research – forecasting for multiple days in the future and the comparison of different LSTM models with ARIMA model – are yet to be widely explored.

This study explores the performance of several LSTM models – vanilla LSTM[4], stacked LSTM[9], bidirectional LSTM[10], convolutional neural network LSTM (CNN LSTM)[11], and ConvLSTM[12] – along with ARIMA model. Moreover, since it is worthwhile to evaluate the performance of these time series forecasting methods for predicting multiple values in the future rather than simply single forecasts, we introduce the concept of k-period performance metrics. These k-period performance metrics extend existing performance metrics used for evaluating single forecasts to the general case of evaluating multiple forecasts in the future. While the proposed concept of k-period performance metrics can be used to extend any performance metric, we specifically choose Mean Absolute Percentage Error (MAPE)[13] and Median Symmetric Accuracy (MdSA)[14] for deriving the k-period performance metrics used in this study.

The main contributions of this study are:

- We evaluate and compare the performance of several LSTM architectures vis-à-vis the ARIMA model in time series analysis and forecasting of country-wise COVID-19 cases for 4 countries.

- We propose and introduce k-period performance metrics for evaluating and comparing the performance of time series forecasting algorithms where forecasts are made for multiple time periods in the future. While this definition can be used for any performance metric in time series forecasting, we extend the definition of MAPE and MdSA metrics to define k-period kMAPE (kMAPE) and k-period MdSA (kMdSA), respectively, in particular.

We also made the code (along with other analyses) publicly available for use at https://github.com/arkobarman/covid-19_timeSeriesAnalysis.

# 3 MATERIALS AND METHODS

## 3.1 Dataset

Time series data including the number of confirmed cases, recovered, and deaths due to COVID-19 for several countries have been made available publicly by researchers at Johns Hopkins University[2] (https://github.com/CSSEGISandData/COVID-19/tree/master/csse_covid_19_data). The dataset consists of time series information about the total daily cumulative number of confirmed cases, recovered, and deaths for almost all countries in the world as well as for individual US (United States of America) states. The data is collected from different sources, such as WHO, the Centers for Disease Control and Prevention (CDC) in the US, and the Chinese Center for Disease Control and Prevention (CCDC) to name a few. In our study, we have used the time series for daily cumulative confirmed COVID-19 cases from 4 countries – US, Italy, Spain, and Germany – from this dataset.

The dataset is updated daily with new information. For our experiments, we have used all data until May 25, 2020. Further, we have truncated the data in the beginning so that all time series points with number of cases less than 100 have been removed. In other words, the first time point in our time series corresponds to the first date when the number of confirmed cases in that particular country was 100 or above. This step has been done to ensure that the noise incorporated in the time series data due to lack of testing in several countries in the initial stages of the disease does not affect the time series data in general.

## 3.2 Time series analysis and forecasting

In this study, we have used several LSTM architectures and ARIMA model for the forecasting of confirmed COVID-19 cases in different countries. These models are briefly described here.

*3.2.1 LSTM Models*

Long short-term memory (LSTM) networks[4] are recurrent neural networks (RNN) widely used in deep learning. LSTMs were designed to process sequences of data and improved upon traditional RNN by using memory cells that can store information in memory for long sequences, and a set of gates to control the flow of this memory information. These innovations allow LSTM to learn longer term dependencies in sequential data.

The following LSTM architectures were used in our experiments:

- Vanilla LSTM: A vanilla LSTM is the simplest LSTM model with a single LSTM unit having multiple neurons. In our experiments, the vanilla LSTM had $n_n = 100$ neurons.
- Stacked LSTM: A stacked LSTM[9] utilizes a stacked architecture of two or more vanilla LSTMs. In our experiments, two vanilla LSTMs with $n_n = 50$ were stacked together.
- Bidirectional LSTM: A bidirectional LSTM[10] architecture splits the neurons of a regular LSTM into two directions – positive time direction for forward states and negative time direction for backward states. As a result, input information from both the past and the future of the current time point is used. In our experiments, bidirectional LSTMs with $n_n = 50$ neurons were used.

- CNN LSTM: Also known as Long-term Recurrent Convolutional Network (LRCN), CNN LSTM[11] architecture incorporates 1-D spatial structure along with temporal structure to learn patterns. Here, 1-D spatial structures are created by grouping together a few time points and considering them as 1-D spatial structures. This unique architecture allows the network to identify patterns in small groups of time points along with the actual temporal patterns in the time series. In our experiments, the CNN LSTM consisted of $n_n = 50$ neurons in the LSTM and $n_f = 64$ convolutional filters along with a max pooling layer in the CNN.
- ConvLSTM: ConvLSTM[12] (Convolutional LSTM) architecture allows convolutions at the gates of the LSTM to capture spatio-temporal patterns. Although it learns spatio-temporal patterns similar to CNN LSTM, the manner in which it does so is different (by using convolutions at the gates). In our experiments, the ConvLSTM used $n_n = 50$ neurons and $n_f = 64$ convolutional filters.

LSTMs are notoriously sensitive to initialization. To tackle the initialization problem, the models were trained on 100 different random initializations and the one with the best (least) validation kMAPE (k-day Mean Absolute Percentage Error) was selected for obtaining the performance on the test set. The data were divided into training, validation, and test sets of size $n_d - 20, 10,$ and $10$ respectively (where $n_d$ is the number of time series sequences obtained from the data).

Further, we trained the LSTMs on varying sizes of input sequences. We used sequence sizes of $n_s = 3, 6, 10, 15$ in our experiments. Adam optimizer with a learning rate of 0.1 was used for training the LSTM models and mean squared error was used as the loss function.

*3.2.2 ARIMA Model*

The Auto-regressive Integrated Moving Average (ARIMA) model is a time series analysis and forecasting method. The ARIMA model has three components:

- Autoregression (AR): An Autoregressive model derives the regression relationship between a changing variable based on its own prior (or lagged) values.
- Integrated (I): Integration refers to modeling the differences between raw observations so that the time series can be considered stationary. In other words, raw values are replaced by the differences between the raw values and the previous values.
- Moving Average (MA): The moving average component in the model incorporates the relationship between an observation and a residual error from a moving average model applied to prior (or lagged) observations.

An ARIMA model has three parameters:

1. $p$, the number of lag observations in the model (or the lag order)
2. $d$, the number of times the raw observations are differenced (or the degree of differencing)
3. $q$, the size of the window for moving average (or the order of moving average)

In our experiments, these parameters were empirically chosen to be the following values: $p = 1, d = 2, q = 2$. Values of the parameters were evaluated using a grid search on each parameter between values 1 and 3, and subsequently chosen empirically to reflect the best performance of the ARIMA model. For all experiments with the ARIMA model, the sequence length was kept as $n_s = 15$.

*3.2.3 k-period Forecasts*

Besides the commonly-used 1-day forecast into the future, we also performed experiments for predicting the time series values for $k$ periods (i.e., $k$ days in this case) in the future. We used $k = 1, 3, 5$ in our experiments. For both the LSTM models and the ARIMA model, the prediction of the 1st day in the future was added to the input sequence for predicting the 2nd day and so on. In general, if the input sequence is $[x_1, x_2, \ldots, x_{n_s-1}, x_{n_s}]$, where the size of the sequence is $n_s$, the forecast for the next day is $\hat{x}_{n_s+1}$. For predicting the subsequent day, the input sequence is modified to $[x_2, x_3, \ldots, x_{n_s-1}, x_{n_s}, \hat{x}_{n_s+1}]$ and the forecast for the 2nd day in the future is $\hat{x}_{n_s+2}$. In this way, forecasting for $k = n_p$ days in the future gives us the k-day predictions as $[\hat{x}_{n_s+1}, \hat{x}_{n_s+2}, \ldots, \hat{x}_{n_s+n_p}]$.

## 3.3 Performance Metrics

For evaluating the performance for time series forecasting models, several performance metrics are commonly used[15]. In this study, we have chosen the Mean Absolute Percentage Error (MAPE) and Median Symmetric Accuracy (MdSA)[14], both of which are measures of error in prediction and expressed as a percentage. MAPE is given by,

$$MAPE = \frac{100}{n} \sum_{i=1}^{n} \frac{|\hat{x}_i - x_i|}{|x_i|} \%  \qquad (1)$$

where $x_i$ denotes the true value at time point $i$, $\hat{x}_i$ denotes the predicted value at time point $i$, and $n$ denotes the number of 1-day predictions made.

MdSA is based on the log accuracy ratio and is given by,

$$MdSA = 100 \times \left\{ \exp\left( Md_{j=1}^n \left( \left| \ln \frac{\hat{x}_j}{x_j} \right| \right) \right) - 1 \right\} \%  \qquad (2)$$

where $Md_{j=1}^n(a_j)$ denotes the median of the numbers $a_j, j = 1, 2, \ldots, n$. MdSA possesses several symmetry and robustness properties[14], thus making it a good choice as a metric in our study. Further, the use of the median operation instead of the most commonly-used mean operation renders this metric immune to the effect of outliers, making it more robust compared to MAPE and other similar metrics.

### 3.3.1 k-period Performance Metrics

Both the described metrics are designed for evaluating multiple 1-period forecasts. In our study, we also seek to predict and evaluate multiple k-period forecasts. Towards this goal, we extend the definition of any performance evaluation metric to incorporate the evaluation of k-period forecasts. Consider $P$ sequences $[x_1^p, x_2^p, \ldots, x_{n_s}^p], p = 1, 2, \ldots, P$, each of length $n_s$. For each of these sequences, we can obtain k-period predictions as described before. These predictions are given by $[\hat{x}_{n_s+1}^p, \hat{x}_{n_s+2}^p, \ldots, \hat{x}_{n_s+k}^p], p = 1, 2, \ldots, P$, each of length $k$.

We define the general k-period metric as,

$$\text{k-period } Metric = \frac{1}{k} \sum_{i=1}^{k} Metric\left( [x_{n_s+i}^1, x_{n_s+i}^2, \ldots, x_{n_s+i}^P], [\hat{x}_{n_s+i}^1, \hat{x}_{n_s+i}^2, \ldots, \hat{x}_{n_s+i}^P] \right) \qquad (3)$$

where $Metric([x^1_{n_s+i}, x^2_{n_s+i}, ..., x^P_{n_s+i}], [\hat{x}^1_{n_s+i}, \hat{x}^2_{n_s+i}, ..., \hat{x}^P_{n_s+i}])$ is any metric which measures the performance of a time series forecast model between the true values ($[x^1_{n_s+i}, x^2_{n_s+i}, ..., x^P_{n_s+i}]$) and the predicted values ($[\hat{x}^1_{n_s+i}, \hat{x}^2_{n_s+i}, ..., \hat{x}^P_{n_s+i}]$). In essence, it is the mean of the performance of prediction over $k$ days.

Specifically, we define the k-period MAPE (kMAPE) as,

$$kMAPE = \frac{1}{k}\sum_{i=1}^{k}\left\{\frac{100}{n}\sum_{p=1}^{P}\left|\frac{\hat{x}^p_{n_s+i} - x^p_{n_s+i}}{x^p_{n_s+i}}\right|\right\}\% \qquad (4)$$

where $x^p_{n_s+i}, i = 1, 2, ..., k$ are true values for $k$ days in the future and $\hat{x}^p_{n_s+i}, i = 1, 2, ..., k$ are the predicted values for $k$ days in the future for the $p$th input sequence.

Similarly, k-period MdSA (kMdSA) is given by,

$$kMdSA = \frac{1}{k}\sum_{i=1}^{k}\left\{100 \times \left[\exp\left(Md^P_{p=1}\left(\left|\ln\frac{\hat{x}^p_{n_s+i}}{x^p_{n_s+i}}\right|\right)\right) - 1\right]\right\} \qquad (5)$$

where the notations are the same as in Eqn. 2 and Eqn. 4.

It is to be noted that in the case of 1-day forecast in the future, the k-day performance metrics are identical to the original definition of the performance metrics. Figure 1 shows how performance metrics are calculated and how they can be extended to define the k-period performance metrics.

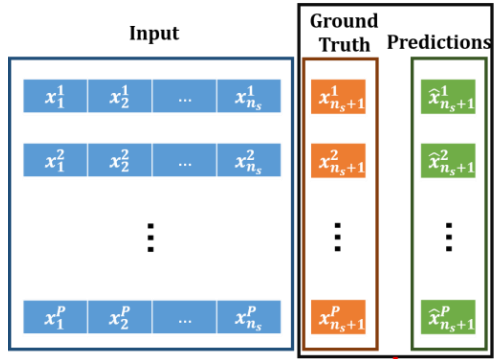

(a) Computing a performance metric

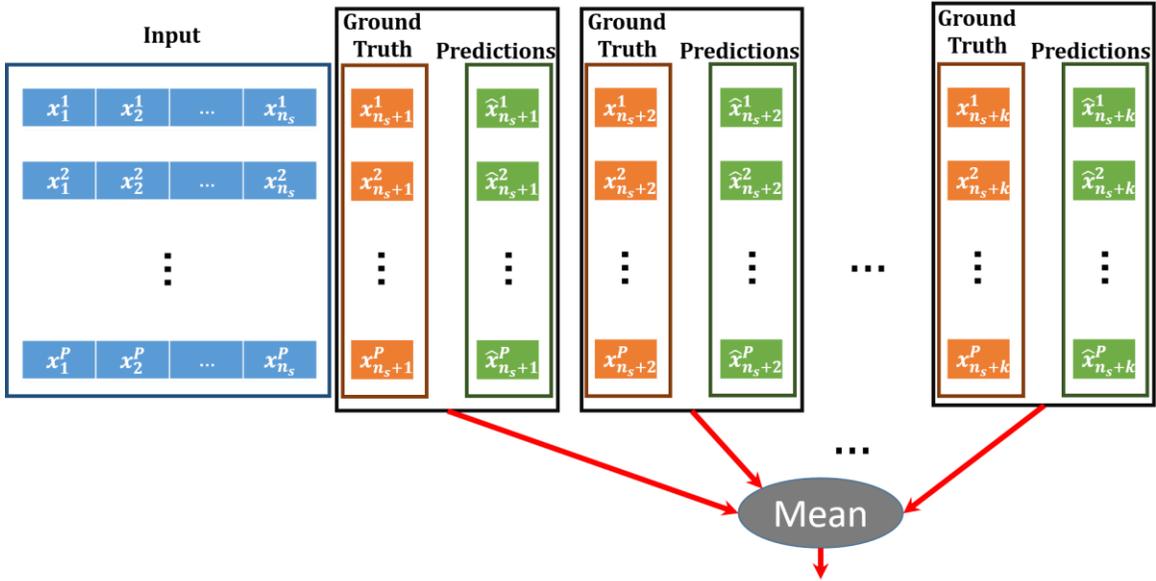

(b) Computing a k-period performance metric

Figure 1. Computing k-period performance metrics by extending the definitions of performance metrics.

# 4 RESULTS

The performance of the different LSTM models and ARIMA models were compared using kMAPE and kMdSA as described. For LSTM models, input sequence length $n_s = 3, 6, 10, 15$ were used in our experiments while for the ARIMA model, only $n_s = 15$ was used.

Table 1 shows the performance of the LSTM models, viz., vanilla LSTM, stacked LSTM, bidirectional LSTM, CNN LSTM, and ConvLSTM, using $n_s = 3$ vis-à-vis the ARIMA model with $n_s = 15$, compared using the kMAPE metric.

| Country | Vanilla LSTM | | | Stacked LSTM | | | Bidirectional LSTM | | | CNN LSTM | | | ConvLSTM | | | ARIMA | | |
|---|---|---|---|---|---|---|---|---|---|---|---|---|---|---|---|---|---|---|
| | $n_p = 1$ | $n_p = 3$ | $n_p = 5$ | $n_p = 1$ | $n_p = 3$ | $n_p = 5$ | $n_p = 1$ | $n_p = 3$ | $n_p = 5$ | $n_p = 1$ | $n_p = 3$ | $n_p = 5$ | $n_p = 1$ | $n_p = 3$ | $n_p = 5$ | $n_p = 1$ | $n_p = 3$ | $n_p = 5$ |
| US | .13 | .58 | .77 | .53 | .47 | 3.4 | .19 | .46 | .66 | .82 | 1.3 | 2.0 | .16 | .38 | .58 | .15 | .41 | .60 |
| IT | .08 | .78 | .58 | .47 | .74 | 1.1 | .32 | .59 | .32 | .62 | .55 | .86 | .07 | .14 | .51 | .07 | .15 | .29 |
| SP | .23 | .51 | .90 | .42 | .63 | .53 | .30 | .75 | .82 | .48 | .64 | .79 | .20 | .32 | .53 | .22 | .29 | .68 |
| GE | .18 | .28 | .40 | .21 | .27 | .24 | .21 | .22 | .21 | .56 | .25 | .64 | .18 | .36 | .46 | .22 | .35 | .45 |

Table 1. k-day Mean Absolute Percentage Error (kMAPE) in forecasting confirmed cases using different time series analysis models. Each of the LSTM models was trained, validated, and tested with number of time series points, $n_s = 3$. For LSTM models with $n_s = 10$ and $n_s = 15$, please see supplementary materials. The ARIMA model used the following parameters: $n_s = 15, (p, d, q) = (1, 2, 2)$. The number of points predicted in the future was set to $n_p = 1, 3$, or $5$. All values in the table are in percentages, e.g. .13 in the table means 0.13%. (US = United States of America, IT = Italy, SP = Spain, GE = Germany)

Table 2 shows the performance of the LSTM models using $n_s = 3$ and the ARIMA model with $n_s = 15$, compared using the kMdSA metric.

| Country | LSTM Models | | | | | | | | | | | | | | | ARIMA | | |
|---|---|---|---|---|---|---|---|---|---|---|---|---|---|---|---|---|---|---|
| | Vanilla LSTM | | | Stacked LSTM | | | Bidirectional LSTM | | | CNN LSTM | | | ConvLSTM | | | | | |
| | $n_p=1$ | $n_p=3$ | $n_p=5$ | $n_p=1$ | $n_p=3$ | $n_p=5$ | $n_p=1$ | $n_p=3$ | $n_p=5$ | $n_p=1$ | $n_p=3$ | $n_p=5$ | $n_p=1$ | $n_p=3$ | $n_p=5$ | $n_p=1$ | $n_p=3$ | $n_p=5$ |
| US | .09 | .64 | .80 | .57 | .47 | 3.4 | .16 | .43 | .67 | .86 | 1.4 | 2.1 | .18 | .38 | .44 | .13 | .39 | .60 |
| IT | .08 | .86 | .65 | .51 | .83 | 1.2 | .31 | .63 | .37 | .62 | .60 | .93 | .09 | .13 | .59 | .06 | .12 | .26 |
| SP | .17 | .54 | .97 | .42 | .70 | .44 | .27 | .75 | .89 | .55 | .66 | .84 | .19 | .33 | .42 | .15 | .28 | .39 |
| GE | .16 | .23 | .36 | .22 | .23 | .21 | .19 | .19 | .17 | .52 | .22 | .63 | .13 | .29 | .34 | .17 | .25 | .26 |

Table 2. k-day Median Symmetric Accuracy (kMdSA) in forecasting using different time series analysis models. Each of the LSTM models was trained, validated, and tested with number of time series points, $n_s = 3$. For LSTM models with $n_s = 10$ and $n_s = 15$, please see supplementary materials. The ARIMA model used the following parameters: $n_s = 15, (p, d, q) = (1, 2, 2)$. The number of points predicted in the future was set to $n_p = 1, 3$, or 5. All values in the table are in percentages, e.g., .09 in the table means 0.09%. (US = United States of America, IT = Italy, SP = Spain, GE = Germany)

The performance of the LSTM models with $n_s = 6$ and the ARIMA model with $n_s = 15$ are compared using the kMAPE metric in Table 3. The values for ARIMA here are identical to the ones shown in Table 1 as they have the same $n_s = 15$ but are repeated here for ease of comparison.

| Country | LSTM Models | | | | | | | | | | | | | | | ARIMA | | |
|---|---|---|---|---|---|---|---|---|---|---|---|---|---|---|---|---|---|---|
| | Vanilla LSTM | | | Stacked LSTM | | | Bidirectional LSTM | | | CNN LSTM | | | ConvLSTM | | | | | |
| | $n_p=1$ | $n_p=3$ | $n_p=5$ | $n_p=1$ | $n_p=3$ | $n_p=5$ | $n_p=1$ | $n_p=3$ | $n_p=5$ | $n_p=1$ | $n_p=3$ | $n_p=5$ | $n_p=1$ | $n_p=3$ | $n_p=5$ | $n_p=1$ | $n_p=3$ | $n_p=5$ |
| US | .07 | .61 | .61 | .70 | 1.1 | .61 | .23 | .58 | 1.6 | .19 | .27 | .81 | .16 | .54 | 1.0 | .15 | .41 | .60 |
| IT | .10 | .10 | .20 | .30 | .10 | .64 | .07 | .32 | .40 | .06 | .16 | .32 | .07 | .10 | .29 | .07 | .15 | .29 |
| SP | .47 | .41 | .60 | .68 | .38 | 1.2 | .20 | .71 | .61 | .61 | .77 | .90 | .20 | .65 | .98 | .22 | .29 | .68 |
| GE | .12 | .33 | .08 | .20 | .36 | 1.2 | .19 | .25 | .73 | .16 | .32 | .86 | .25 | .32 | .71 | .22 | .35 | .45 |

Table 3. k-day Mean Absolute Percentage Error (kMAPE) in forecasting confirmed cases using different time series analysis models. Each of the LSTM models was trained, validated, and tested with number of time series points, $n_s = 6$. For LSTM models with $n_s = 10$ and $n_s = 15$, please see supplementary materials. The ARIMA model used the following parameters: $n_s = 15, (p, d, q) = (1, 2, 2)$. The number of points predicted in the future was set to $n_p = 1, 3,$ or 5. All values in the table are in percentages, e.g., .07 in the table means 0.07%. (US = United States of America, IT = Italy, SP = Spain, GE = Germany)

Table 4 shows the comparison of the performance of the LSTM models with $n_s = 6$ and the ARIMA model with $n_s = 15$ using the kMdSA metric. The values for ARIMA are repeated in Table 4 from Table 2 for ease of comparison.

| Country | LSTM Models | | | | | | | | | | | | | | | ARIMA | | |
|---|---|---|---|---|---|---|---|---|---|---|---|---|---|---|---|---|---|---|
| | Vanilla LSTM | | | Stacked LSTM | | | Bidirectional LSTM | | | CNN LSTM | | | ConvLSTM | | | | | |
| | $n_p = 1$ | $n_p = 3$ | $n_p = 5$ | $n_p = 1$ | $n_p = 3$ | $n_p = 5$ | $n_p = 1$ | $n_p = 3$ | $n_p = 5$ | $n_p = 1$ | $n_p = 3$ | $n_p = 5$ | $n_p = 1$ | $n_p = 3$ | $n_p = 5$ | $n_p = 1$ | $n_p = 3$ | $n_p = 5$ |
| US | .05 | .60 | .56 | .75 | 1.2 | .57 | .25 | .53 | 1.6 | .16 | .25 | .88 | .16 | .54 | 1.1 | .13 | .39 | .60 |
| IT | .09 | .10 | .17 | .29 | .07 | .66 | .07 | .33 | .42 | .05 | .17 | .24 | .08 | .09 | .23 | .06 | .12 | .26 |
| SP | .51 | .45 | .62 | .72 | .38 | 1.2 | .12 | .76 | .60 | .65 | .86 | 1.0 | .11 | .66 | 1.0 | .15 | .28 | .39 |
| GE | .05 | .26 | .06 | .15 | .32 | 1.2 | .18 | .20 | .70 | .10 | .28 | .84 | .09 | .27 | .69 | .17 | .25 | .26 |

Table 4. k-day Median Symmetric Accuracy (kMdSA) in forecasting using different time series analysis models. Each of the LSTM models was trained, validated, and tested with number of time series points, $n_s = 6$. For LSTM models with $n_s = 10$ and $n_s = 15$, please see supplementary materials. The ARIMA model used the following parameters: $n_s = 15, (p, d, q) = (1, 2, 2)$. The number of points predicted in the future was set to $n_p = 1, 3,$ or 5. All values in the table are in percentages, e.g., .05 in the table means 0.05%. (US = United States of America, IT = Italy, SP = Spain, GE = Germany)

Figure 2 shows example forecasts for all models vis-à-vis the ground truth using sequence length, $n_s = 6$ for the LSTM models and $n_s = 15$ for ARIMA model. Predictions were made for 3 days, i.e. $n_p = 6$ (May 23 to May 25).

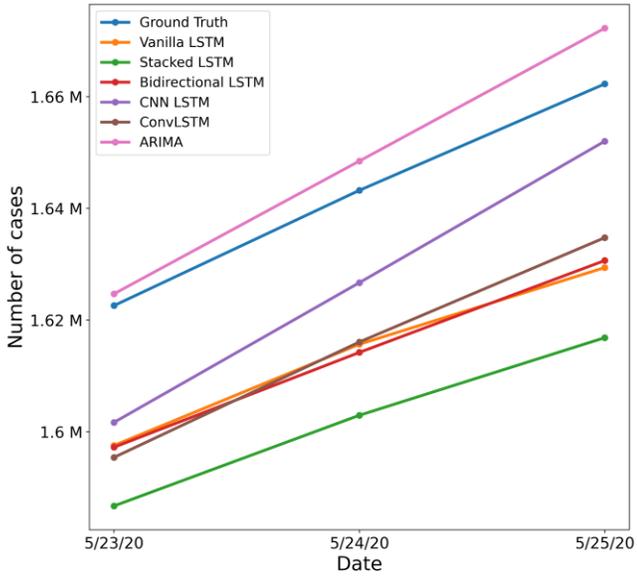

(a) United States

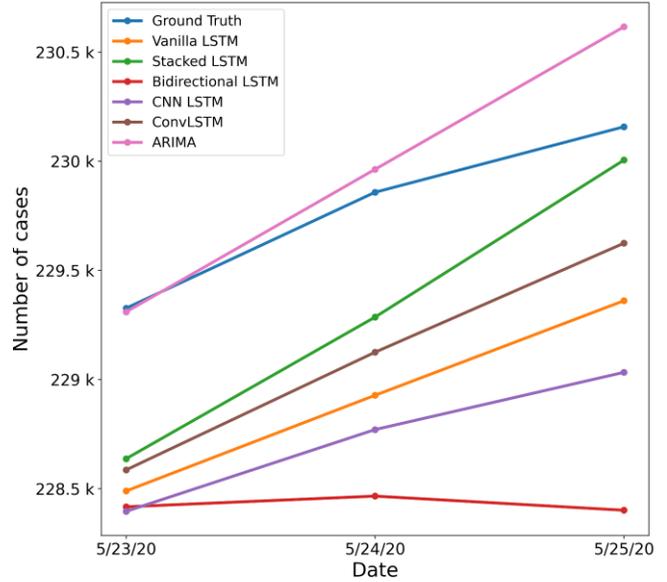

(b) Italy

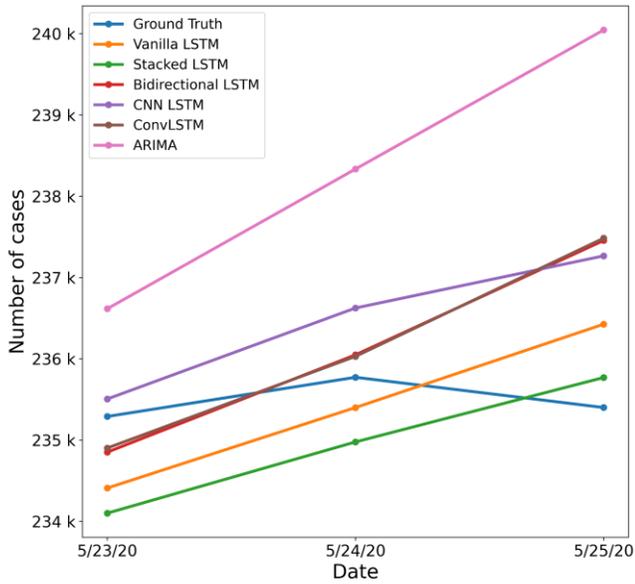

(c) Spain

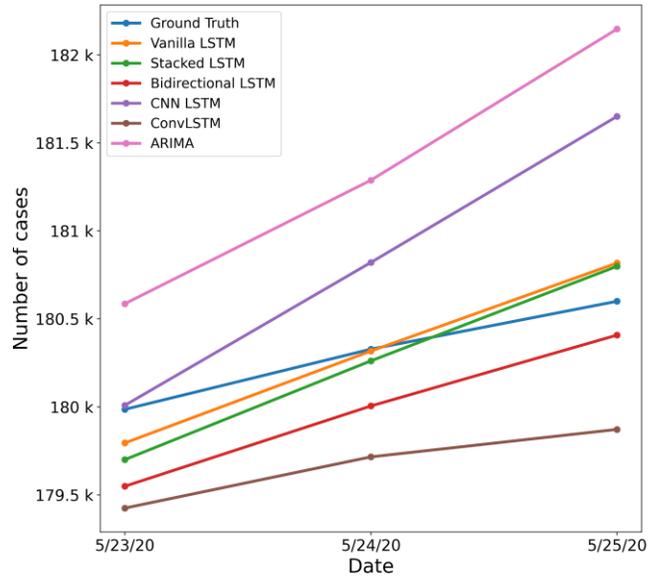

(d) Germany

Figure 2. Sample forecasts of all models compared with the ground truth numbers for $n_p = 3$ days (May 23 to May 25) using sequence length of $n_s = 6$ for the LSTM models and $n_s = 15$ for the ARIMA model.

# 5 DISCUSSION

Short- and long-term forecasts and predictions of the number of cases in any epidemic or pandemic can be a useful tool for several government agencies and policymakers in different countries to prepare for the progression and spread of a disease. Given the rapid growth of COVID-19 cases worldwide, it is imperative that these tools be widely available to relevant agencies and personnel to aid in preparation (e.g. number of hospital beds and ventilators, requirement and availability of healthcare workers, and the demand for personal protective equipment (PPE)) which would be directly reflected in better outcomes for patients as well as the number of lives saved due to a deadly or fatal disease.

In this work, we have compared the ARIMA model with variants of a supervised model (LSTM) for forecasting country-wise COVID-19 cases for 4 different countries – United States of America, Italy, Spain, and Germany. Similar analyses can be performed with the data from any country and even at more granular levels, such as states, counties, cities, and districts to obtain an insight into the preparation required to combat the spread of the disease.

Our results indicate ARIMA performs reasonably well and is often superior to the variants of the supervised LSTM models used in this study. This makes the ARIMA model suitable for baseline performance comparisons and can be used for developing better models for time series analysis and forecasting.

As expected, the performance of all the models deteriorates slightly as the number of forecast periods (i.e., days in our case) in the future is increased. This drop in performance is attributed to the fact that the prediction for the 2$^{nd}$ day uses the prediction for the 1$^{st}$ day in its input sequence, and so on. As a result, the errors in prediction are propagated as the forecasts are made for more and more days into the future. This is the reason why time series forecasting models presented in this study are unsuitable for long-term forecasting.

It is also interesting to note that the performance of the vanilla LSTM is comparable and often better than the performance of the more complex LSTM models discussed here. This pattern indicates that more complex models are likely not necessary for this forecasting problem.

Since we have used only data starting with the day when 100 cases were reported for a country, the number of training sequences for each country has been different. Given the nature of deep learning models such as LSTM, improvements in performance often correlate with greater number of training data. As a result, it can be seen how the forecast performance of the LSTM models for US, Italy, and Germany are better than those for Spain since the number of training samples for Spain were fewer compared to the other countries. It is expected that the performance of all the LSTM models would improve over time as more and more data becomes available for all countries worldwide, thus making more training data available. The ARIMA model is, however, immune to these problems.

In light of the training set size, the choice of models for forecasting becomes obvious. If there are only a small number of training samples available, the ARIMA model can be chosen for forecasting. However, once more and more data is available for training the LSTM models, the best among these models can be chosen if its performance supersedes that of the ARIMA model.

On the other hand, whereas LSTM models in our experiment were fairly stable even for very short input sequences of length, $n_s = 3$ and $n_s = 6$, the sequence length for the ARIMA model is kept at $n_s = 15$, since a short sequence length is deemed to be insufficient for exploring the correlation in the sequence by ARIMA. This makes LSTM models more attractive if we are dealing with sequences of shorter length.

From Figure 2, it is also apparent that ARIMA models slightly overestimate the numbers while LSTM models are more likely to underestimate in their forecasts when compared to the actual ground truth data.

There are some limitations to this work. Out of several possible LSTM models, only five were evaluated in this study. Moreover, there is a plethora of time series analysis and forecasting algorithms that can be evaluated for this particular problem. The time series analysis and forecasting presented in this study are restricted to short term forecasts only. Forecasting over longer terms would require more parameters to be considered in the modeling that is beyond the scope of this work.

The data that has been made available by the public agencies of different countries might be prone to errors and the limited testing capabilities for COVID-19 in certain countries are likely to affect the results of the algorithms, thus yielding lower estimates. Moreover, the availability of testing and rate of infection within a country might vary between different demographics due to economic conditions, the cost of testing, and the availability of insurance among the different demographics[16]. The present study also does not model the changes in the public health policy of different countries which would possibly affect the future estimates for the numbers of patients suffering from COVID-19. However, it might be noted that these changes in public health policy would most likely affect long term forecasts rather than short term trends.

## 6 CONCLUSION

COVID-19 is a global public health crisis. Forecasts and predictions of cases for the future are essential for government agencies and policymakers to combat the situation by planning for the needs of healthcare personnel and the availability of healthcare equipment and PPEs. In this study, we have presented the time series analysis and forecasting of the number of COVID-19 cases in 4 countries – United States, Italy, Spain, and Germany – using the ARIMA model and several LSTM architectures. We have also proposed the concept of k-period performance metrics designed to evaluate the performance of a time series analysis model that forecasts for multiple periods in the future. In particular, we have extended the definition of MAPE and MdSA to define k-period MAPE (kMAPE) and k-period MdSA (kMdSA). Our results indicate that LSTM models perform comparably with the ARIMA model. However, both models have their advantages and disadvantages, and the choice of models is dictated by the availability of training samples as well as the sequence length. Further research would include the development and evaluation of other time series forecasting models as well as the evaluation of these models as more data becomes available.

## References

1    WHO Director-General's opening remarks at the media briefing on COVID-19 - 11 March 2020. https://www.who.int/dg/speeches/detail/who-director-general-s-opening-remarks-at-the-media-briefing-on-covid-19---11-march-2020 (accessed 4 Jun 2020).

2    Dong E, Du H, Gardner L. An interactive web-based dashboard to track COVID-19 in real time. *The Lancet Infectious Diseases* 2020;**20**:533–4. doi:10.1016/S1473-3099(20)30120-1

3    Liu Y, Gayle AA, Wilder-Smith A, *et al.* The reproductive number of COVID-19 is higher compared to SARS coronavirus. *J Travel Med* 2020;**27**. doi:10.1093/jtm/taaa021

4    Hochreiter S, Schmidhuber J. Long Short-Term Memory. *Neural Comput* 1997;**9**:1735–1780. doi:10.1162/neco.1997.9.8.1735

5    Nelson BK. Time Series Analysis Using Autoregressive Integrated Moving Average (ARIMA) Models. *Academic Emergency Medicine* 1998;**5**:739–44. doi:10.1111/j.1553-2712.1998.tb02493.x

6    Bayyurt L, Bayyurt B. Forecasting of COVID-19 Cases and Deaths Using ARIMA Models. *medRxiv* 2020;:2020.04.17.20069237. doi:10.1101/2020.04.17.20069237

7    Benvenuto D, Giovanetti M, Vassallo L, *et al.* Application of the ARIMA model on the COVID-2019 epidemic dataset. *Data in Brief* 2020;**29**:105340. doi:10.1016/j.dib.2020.105340

8    Chintalapudi N, Battineni G, Amenta F. COVID-19 virus outbreak forecasting of registered and recovered cases after sixty day lockdown in Italy: A data driven model approach. *Journal of Microbiology, Immunology and Infection* 2020;**53**:396–403. doi:10.1016/j.jmii.2020.04.004

9    Graves A, Mohamed A, Hinton G. Speech Recognition with Deep Recurrent Neural Networks. *arXiv:13035778 [cs]* Published Online First: 22 March 2013.http://arxiv.org/abs/1303.5778 (accessed 4 Jun 2020).

10   Graves A, Schmidhuber J. Framewise phoneme classification with bidirectional LSTM and other neural network architectures. *Neural Networks* 2005;**18**:602–10. doi:10.1016/j.neunet.2005.06.042

11   Wang J, Yu L-C, Lai KR, *et al.* Dimensional Sentiment Analysis Using a Regional CNN-LSTM Model. In: *Proceedings of the 54th Annual Meeting of the Association for Computational Linguistics (Volume 2: Short Papers)*. Berlin, Germany: : Association for Computational Linguistics 2016. 225–230. doi:10.18653/v1/P16-2037

12   SHI X, Chen Z, Wang H, *et al.* Convolutional LSTM Network: A Machine Learning Approach for Precipitation Nowcasting. In: Cortes C, Lawrence ND, Lee DD, *et al.*, eds. *Advances in Neural Information Processing Systems 28*. Curran Associates, Inc. 2015. 802–810.http://papers.nips.cc/paper/5955-convolutional-lstm-network-a-machine-learning-approach-for-precipitation-nowcasting.pdf (accessed 4 Jun 2020).

13   Armstrong JS, Carbone R. Evaluation of Extrapolative Forecasting Methods: Results of a Survey of Academicians and Practitioners. Rochester, NY: : Social Science Research Network 2005. https://papers.ssrn.com/abstract=663667 (accessed 4 Jun 2020).

14   Morley SK, Brito TV, Welling DT. Measures of Model Performance Based On the Log Accuracy Ratio. *Space Weather* 2018;**16**:69–88. doi:10.1002/2017SW001669

15   Botchkarev A. Performance Metrics (Error Measures) in Machine Learning Regression, Forecasting and Prognostics: Properties and Typology. *IJIKM* 2019;**14**:045–76. doi:10.28945/4184

16   Yancy CW. COVID-19 and African Americans. *JAMA* 2020;**323**:1891–2. doi:10.1001/jama.2020.6548

# Supplementary Material

Tables S1 and S2 show the performance of the different LSTM models with $n_s = 10$ and the ARIMA model with $n_s = 15$ using the kMAPE and the kMdSA metrics respectively.

| Country | Vanilla LSTM | | | Stacked LSTM | | | Bidirectional LSTM | | | CNN LSTM | | | ConvLSTM | | | ARIMA | | |
|---|---|---|---|---|---|---|---|---|---|---|---|---|---|---|---|---|---|---|
| | $n_p=1$ | $n_p=3$ | $n_p=5$ | $n_p=1$ | $n_p=3$ | $n_p=5$ | $n_p=1$ | $n_p=3$ | $n_p=5$ | $n_p=1$ | $n_p=3$ | $n_p=5$ | $n_p=1$ | $n_p=3$ | $n_p=5$ | $n_p=1$ | $n_p=3$ | $n_p=5$ |
| US | .23 | .48 | .97 | .25 | .48 | 1.9 | .30 | .81 | .69 | .55 | .90 | .42 | .13 | .58 | .52 | .15 | .41 | .60 |
| IT | .05 | .50 | .21 | .30 | .34 | .68 | .11 | .15 | .09 | .15 | .25 | .60 | .14 | .31 | .17 | .07 | .15 | .29 |
| SP | .23 | .73 | 1.2 | 1.1 | .41 | .93 | .23 | 1.5 | .49 | .70 | .92 | .78 | .20 | .55 | .70 | .22 | .29 | .68 |
| GE | .14 | .24 | .27 | .23 | .45 | .41 | .16 | .27 | .25 | .19 | .24 | .89 | .12 | .25 | .33 | .22 | .35 | .45 |

Table S1. k-day Mean Absolute Percentage Error (kMAPE) in forecasting confirmed cases using different time series analysis models. Each of the LSTM models was trained, validated, and tested with number of time series points, $n_s = 10$. The ARIMA model used the following parameters: $n_s = 15, (p, d, q) = (1, 2, 2)$. The number of points predicted in the future was set to $n_p = 1, 3,$ or 5. All values in the table are in percentages, e.g. .23 in the table means 0.23%. (US = United States of America, IT = Italy, SP = Spain, GE = Germany)

| Country | Vanilla LSTM | | | Stacked LSTM | | | Bidirectional LSTM | | | CNN LSTM | | | ConvLSTM | | | ARIMA | | |
|---|---|---|---|---|---|---|---|---|---|---|---|---|---|---|---|---|---|---|
| | $n_p=1$ | $n_p=3$ | $n_p=5$ | $n_p=1$ | $n_p=3$ | $n_p=5$ | $n_p=1$ | $n_p=3$ | $n_p=5$ | $n_p=1$ | $n_p=3$ | $n_p=5$ | $n_p=1$ | $n_p=3$ | $n_p=5$ | $n_p=1$ | $n_p=3$ | $n_p=5$ |
| US | .21 | .51 | 1.1 | .21 | .48 | 2.0 | .31 | .84 | .75 | .57 | .90 | .40 | .08 | .57 | .61 | .13 | .39 | .60 |
| IT | .05 | .57 | .22 | .34 | .37 | .69 | .13 | .15 | .09 | .16 | .26 | .59 | .15 | .31 | .14 | .06 | .12 | .26 |
| SP | .22 | .76 | 1.2 | 1.2 | .43 | .98 | .19 | 1.6 | .48 | .73 | .98 | .79 | .16 | .56 | .74 | .15 | .28 | .39 |
| GE | .12 | .20 | .27 | .24 | .43 | .42 | .11 | .24 | .20 | .17 | .16 | .91 | .06 | .21 | .34 | .17 | .25 | .26 |

Table S2. k-day Median Symmetric Accuracy (kMdSA) in forecasting confirmed cases using different time series analysis models. Each of the LSTM models was trained, validated, and tested with number of time series points, $n_s = 10$. The ARIMA model used the following parameters: $n_s = 15, (p, d, q) = (1, 2, 2)$. The number of points predicted in the future was set to $n_p = 1, 3,$ or 5. All values in the table are in percentages, e.g., .21 in the table means 0.21%. (US = United States of America, IT = Italy, SP = Spain, GE = Germany)

Tables S3 and S4 compare the performance of the 5 different LSTM models with $n_s = 15$ and the ARIMA model with $n_s = 15$ using the kMAPE and the kMdSA metrics respectively.

| Country | Vanilla LSTM | | | Stacked LSTM | | | Bidirectional LSTM | | | CNN LSTM | | | ConvLSTM | | | ARIMA | | |
|---|---|---|---|---|---|---|---|---|---|---|---|---|---|---|---|---|---|---|
| | $n_p=1$ | $n_p=3$ | $n_p=5$ | $n_p=1$ | $n_p=3$ | $n_p=5$ | $n_p=1$ | $n_p=3$ | $n_p=5$ | $n_p=1$ | $n_p=3$ | $n_p=5$ | $n_p=1$ | $n_p=3$ | $n_p=5$ | $n_p=1$ | $n_p=3$ | $n_p=5$ |
| US | .31 | .68 | 3.8 | .24 | .35 | 5.1 | .16 | .97 | 1.0 | 6.0 | 1.6 | 1.7 | .16 | 1.7 | .32 | .15 | .41 | .60 |
| IT | .25 | .21 | .27 | .18 | .23 | 1.9 | .12 | .47 | .49 | .46 | .28 | .58 | .07 | .18 | .55 | .07 | .15 | .29 |
| SP | .21 | .86 | 1.0 | 1.5 | .32 | 2.7 | .19 | .47 | .50 | 1.1 | .55 | .69 | .24 | .37 | .44 | .22 | .29 | .68 |
| GE | .18 | .14 | .73 | .22 | .40 | .28 | .14 | .14 | .15 | .21 | .56 | .39 | .20 | .63 | .39 | .22 | .35 | .45 |

Table S3. k-day Mean Absolute Percentage Error (kMAPE) in forecasting confirmed cases using different time series analysis models. Each of the LSTM models was trained, validated, and tested with number of time series points, $n_s = 15$. The ARIMA model used the following parameters: $n_s = 15, (p, d, q) = (1, 2, 2)$. The number of points predicted in the future was set to $n_p = 1, 3,$ or 5. All values in the table are in percentages, e.g., .31 in the table means 0.31%. (US = United States of America, IT = Italy, SP = Spain, GE = Germany)

| Country | Vanilla LSTM | | | Stacked LSTM | | | Bidirectional LSTM | | | CNN LSTM | | | ConvLSTM | | | ARIMA | | |
|---|---|---|---|---|---|---|---|---|---|---|---|---|---|---|---|---|---|---|
| | $n_p=1$ | $n_p=3$ | $n_p=5$ | $n_p=1$ | $n_p=3$ | $n_p=5$ | $n_p=1$ | $n_p=3$ | $n_p=5$ | $n_p=1$ | $n_p=3$ | $n_p=5$ | $n_p=1$ | $n_p=3$ | $n_p=5$ | $n_p=1$ | $n_p=3$ | $n_p=5$ |
| US | .38 | .70 | 4.0 | .22 | .31 | 5.6 | .13 | 1.1 | 1.1 | 6.0 | 1.8 | 1.5 | .12 | 1.5 | .29 | .13 | .39 | .60 |
| IT | .30 | .20 | .21 | .21 | .23 | 2.0 | .13 | .47 | .49 | .44 | .27 | .58 | .06 | .17 | .63 | .06 | .12 | .26 |
| SP | .13 | .96 | .97 | 1.4 | .36 | 2.8 | .11 | .48 | .54 | 1.1 | .54 | .72 | .20 | .42 | .38 | .15 | .28 | .39 |
| GE | .18 | .11 | .70 | .21 | .35 | .26 | .15 | .11 | .13 | .15 | .59 | .37 | .16 | .61 | .41 | .17 | .25 | .26 |

Table S4. k-day Median Symmetric Accuracy (kMdSA) in forecasting confirmed cases using different time series analysis models. Each of the LSTM models was trained, validated, and tested with number of time series points, $n_s = 15$. The ARIMA model used the following parameters: $n_s = 15, (p, d, q) = (1, 2, 2)$. The number of points predicted in the future was set to $n_p = 1, 3,$ or 5. All values in the table are in percentages, e.g., 0.38 in the table means 0.38%. (US = United States of America, IT = Italy, SP = Spain, GE = Germany)